\documentclass[runningheads]{llncs}
\usepackage{amssymb}

\usepackage[ruled,vlined,linesnumbered]{algorithm2e} 
\usepackage{amsmath}
\usepackage{amssymb}
\usepackage{booktabs}
\usepackage{multirow}
\usepackage{xcolor}
\usepackage{colortbl}
\usepackage{graphicx}
\usepackage{marvosym}
\usepackage[colorlinks=true, 
            linkcolor=red,
            citecolor=blue,
            urlcolor=blue,
            linktoc=all]{hyperref}
\usepackage[T1]{fontenc}

\usepackage{graphicx,verbatim}
\begin{document}
\title{Mask to Concept: Auto-Promptable SAM3 via Efficient Test-Time Concept Embedding Search for Few-Shot Annotation}
%

\author{Quan Zhou$^{\dag}$\inst{1} 
\and Shaoqing Zhai$^{\dag}$\inst{1} 
\and Qiang Hu\textsuperscript{(\Letter)}\inst{2} 
\and Jia Chen\inst{3} 
\and Qiang Li\inst{2} \\ 
\and Zhiwei Wang\textsuperscript{(\Letter)}\inst{2}} 
\titlerunning{Auto-promptable SAM3 via Concept Embedding Searching}
\authorrunning{Q. Zhou et al.}
\institute{Wuhan University of Technology \and  Huazhong University of Science and Technology \and Changzhou United lmaging Healthcare Surgical Technology Co. Ltd.\\
\email{\{zhouquan910, zhaishaoqing\}@whut.edu.cn, \{huqiang77, zwwang\}@hust.edu.cn}}

\maketitle              

\def\thefootnote{$\dag$}\footnotetext{Equal~contribution; \textrm{\Letter}~Corresponding author.}

\begin{abstract}
Transforming foundation segmentation models from human-prompted tools into auto-promptable annotators is critical for scalable medical data annotation. Current methods commonly depend on external feature matchers or auxiliary networks to automate geometric prompting, but introducing architectural overhead and limiting performance scalability. Although SAM3 natively supports concept segmentation via reusable text prompts, its direct use in medical imaging is hindered by a lack of fine-grained clinical knowledge and the ambiguity of human-written descriptions. In this work, we propose Mask to Concept (M2C), an efficient framework that adapts SAM3 for medical few-shot annotation \emph{without} external modules, parameter retraining, or manual text engineering. Using only a few labeled images, M2C enables SAM3 to automatically search for transferable visual concepts entirely within its frozen architecture: it initializes a learnable concept embedding, uses it to prompt segmentation, and updates the embedding by gradients of minimizing the concept segmentation error. We further introduce a Hybrid Uncertainty Estimation (HUE) module that calculates the prediction entropy and maps concept predictions back to the box prompts, measuring concept-geometry prompting inconsistency. Highly uncertain samples are flagged actively for human correction, and the corrected masks are then fed back to M2C to continuously search for more precise concept embeddings, forming a self-enhancing annotation loop with minimal expert effort. Experiments on medical segmentation benchmarks show that our method achieves SOTA few-shot segmentation performance and outstanding annotation efficiency, offering a practical and efficient pathway toward scalable medical image labeling. Codes are at \href{https://github.com/Huster-Hq/M2C}{https://github.com/Huster-Hq/M2C}.


\end{abstract}
\section{Introduction}
High-quality, large-scale annotations are essential for modern medical AI, yet obtaining pixel-level segmentation labels is prohibitively expensive \cite{tajbakhsh2020embracing,hu2024sali}.
This annotation bottleneck has motivated research into training-free tools. Models like SAM \cite{kirillov2023segment,ravi2024sam} and its clinical variants (\textit{e.g.}, MedSAM~\cite{ma2024segment}) offer a promising start, enabling segmentation from simple geometric prompts. However, they operate as per-image interactive assistants, requiring manual input for each case and limiting scalability for large-scale annotation.

To amortize human effort, recent works seek to develop automatic annotators driven by a few labeled samples.
These works can be categorized into two types:
The first category is \textit{training-based methods}~\cite{butoi2023universeg,rakic2024tyche,wong2025multiverseg,cheng2024few,hu2026samix}, which heavily rely on pre-training under the few-shot segmentation (FSS) paradigm on large-scale medical datasets.
However, their performance tends to be constrained within the training data distribution, exhibiting poor generalization and flexibility.
Conversely, \textit{training-free methods} typically leverage SAM by introducing automated geometric prompt generation mechanism to drive the annotation process.
However, they rely on external modules, such as separate feature matching modules \cite{liu2023matcher,ayzenberg2024protosam,zhu2025maup,zhang2023personalize} or auxiliary networks \cite{wu2023self,liu2025synpo,hu2025first}, to establish cross-instance correspondence.
This not only complicates the pipeline but often fails to fully leverage SAM's internal reasoning, resulting in inconsistent quality: performance lags behind per-image manual prompting and rarely improves with more references.

The release of SAM3 \cite{carion2025sam}, with its native support for promptable concept segmentation using texts, marks a pivotal shift. By moving from case-specific geometric prompts to reusable semantic concepts carried by text prompts, it opens a direct pathway toward a concept-driven annotation assistant. This raises our core research question:
\emph{Can we unlock SAM3's concept segmentation to build a cross-instance few-shot annotator entirely within its native capabilities?}
However, achieving this in medical imaging is challenging. SAM3 lacks inherent knowledge of fine-grained clinical concepts (\textit{e.g.}, specific lesion subtypes), and human-written text prompts are often too vague for reliable \cite{zhou2022learning,gal2022image}, automated labeling across a diverse dataset.

To address these challenges, we make the following major contributions:
\textbf{(1)} We introduce Mask to Concept (M2C), a novel mechanism that enables SAM3 to automatically search for medical visual concepts from annotated masks, bypassing ambiguous text prompts and adapting to the medical domain without FSS-Pretraining and SAM3-retraning.
Instead of manual prompt engineering, M2C initializes an efficient \emph{learnable} concept embedding and refines it through a differentiable pathway: given a few annotated images, the embedding prompts SAM3 to predict masks, and the prediction error is backpropagated to update the embedding. This process is fast, memory-efficient, and grows more accurate with more examples, naturally supporting incremental few-shot learning.
\textbf{(2)} We further integrate M2C into a human-in-the-loop annotation system, where the M2C embedding enables SAM3 to annotate the rest unlabeled data autonomously.
To identify samples requiring human verification, we design a hybrid uncertainty estimation (HUE) module to evaluate prediction uncertainty. Low-uncertainty samples are auto-labeled, while high-uncertainty ones are sent for human verification.
Newly corrected masks then refine the concept embedding, closing an active learning loop that continuously improves annotation quality with minimal human effort.
\textbf{(3)} Extensive experiments demonstrate that M2C achieves SOTA FSS performance: under the one-shot setting, M2C outperforms existing methods by at least $4.2\%$ and $11.8\%$ in Dice on Kvasir-SEG and ISIC-2017, respectively.
Moreover, the proposed annotation system exhibits superior efficiency and data generalization on both datasets, showing its practical significance.

\section{Method}
\label{sec:method}
\subsection{Overview}
\label{subsec:overview}
\begin{figure}[t]
    \centering
    \includegraphics[width=\textwidth]{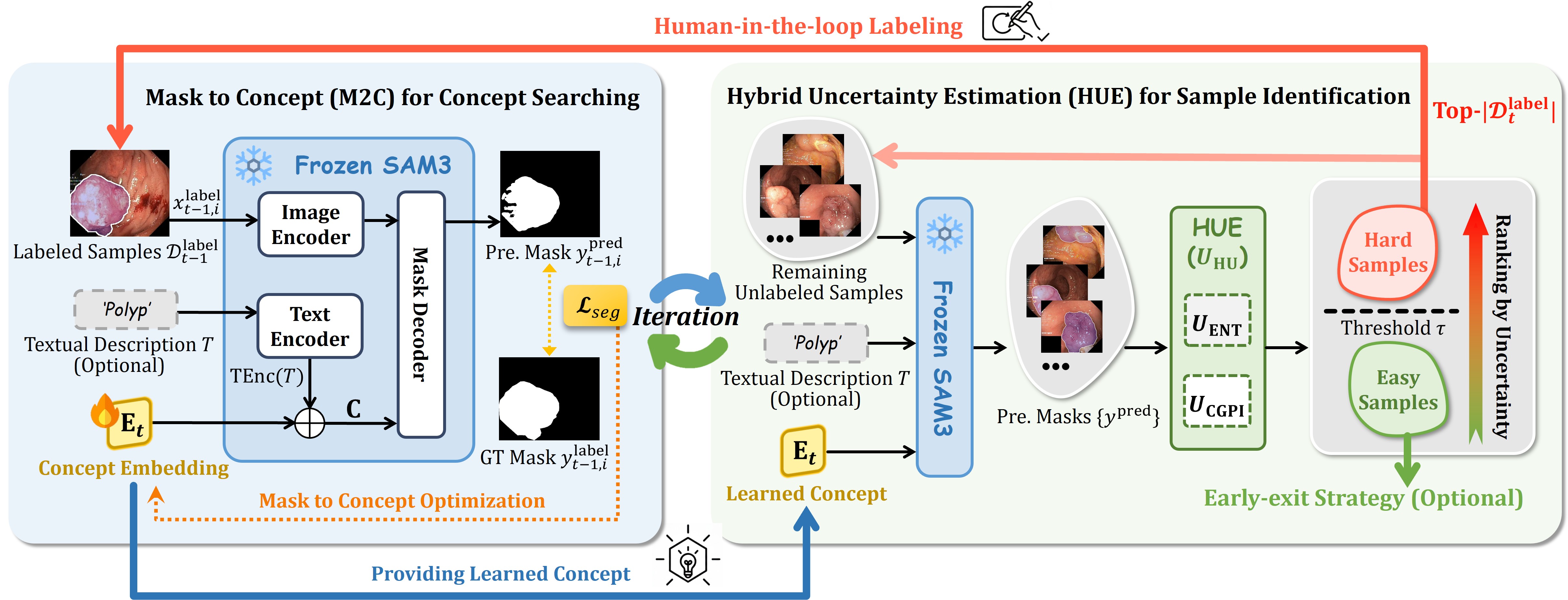}
    \vspace{-0.2cm}
    \caption{Overview of the proposed human-in-the-loop annotation framework. It consists of two core iterative modules: Mask to Concept (M2C) and Hybrid Uncertainty Estimation (HUE). During each iteration, given a few human-labeled samples from the previous iteration, M2C employs a continuous searching on the dataset-specific concept embedding to automatically segment the remaining unlabeled samples. Following this, HUE estimates the uncertainties of these predictions and applies tailored strategies to the samples based on the uncertainty as they transition into the next iteration.}
    \vspace{-0.4cm}
\label{fig1}
\end{figure}

As illustrated in Fig.~\ref{fig1}, our human-in-the-loop annotation framework based on SAM3 is empowered by two iterative modules: Mask to Concept (M2C) and Hybrid Uncertainty Estimation (HUE).
Given a specific medical dataset, users first annotate the masks for one or several samples using SAM3 with simple geometric prompts (\textit{e.g.}, boxes or points).
Then M2C rapidly learns a \textbf{\textit{concept embedding}} specific to the dataset derived from these mask annotations, allowing SAM3 to automatically segment the remaining unlabeled samples via the concept prompting mechanism.
Next, HUE evaluates these predicted masks and ranks samples based on the estimated uncertainty scores.
High-uncertainty samples will be proactively selected for manual label correction, with the corrected labels fed back to M2C for continuous search of the concept embedding.
Low-uncertainty samples can be optionally verified and removed from the unlabeled sample pool.
These two modules are iterated until the entire dataset is annotated, enabling scalable accurate annotation with cost-effective labor costs.

\subsection{Mask-to-Concept for Concept Searching}
To overcome SAM3's inherent reliance on text descriptions for concept segmentation and enable its flexible adaptation to arbitrary scenarios, we propose Mask-to-Concept (M2C). M2C learns a concept embedding from a few labeled samples, transforming SAM3 into a dataset-specific automatic segmenter while keeping its backbone parameters strictly frozen. 

Specifically, given a medical image $x$ and a human-written text prompt $T$ (\textit{e.g.}, a category name or a sentence), SAM3 originally utilizes a text encoder to obtain a concept embedding $\mathtt{TEnc}(T)$, and then predicts a segmentation mask $y^{\text{pred}} = \mathtt{SAM}(x, \mathtt{TEnc}(T))$.
To search a better concept, we introduce a learnable vector $\mathbf{E}$ to revise the original concept embedding, and thus the segmentation process can be re-formulated as $y^{\text{pred}} = \mathtt{SAM}(x, \mathtt{TEnc}(T) + \mathbf{E})$.

With this modification, a continual searching strategy is proposed to iteratively refine $\mathbf{E}$, enabling it to progressively adapt to remaining hard samples.
At the $t$-th iteration, given a set of human-annotated samples $\mathcal{D}_{t-1}^{\text{label}} = \{(x_{t-1,i}^{\text{label}}, y_{t-1,i}^{\text{label}})\}_{i=1}^{|\mathcal{D}_{t-1}^{\text{label}}|}$ from the previous iteration, we freeze the entire SAM3 model and optimize only the learnable embedding.
Here, $x_{t-1,i}^{\text{label}}$ and $y_{t-1,i}^{\text{label}}$ represent the $i$-th image and labeled mask, respectively.
Initialized with $\mathbf{E}_{t-1}$ from the prior iteration, the embedding is optimized end-to-end by minimizing the segmentation loss via gradient descent:
\begin{equation}
\mathbf{E}_{t} = \mathbf{E}_{t-1} - \eta \nabla_{\mathbf{E}_{t-1}} \bigg( \frac{1}{|\mathcal{D}_{t-1}^{\text{label}}|} \sum_{i=1}^{|\mathcal{D}_{t-1}^{\text{label}}|} \mathcal{L}_{\text{seg}} \big( y^{\text{pred}}_{t-1,i}, y_{t-1,i}^{\text{label}} \big) \bigg),
\end{equation}
where $y^{\text{pred}}_{t-1,i} = \mathtt{SAM}(x_{t-1,i}^{\text{label}}, \mathtt{TEnc(T) + \mathbf{E}_{t-1}})$, and $\eta$ is the searching step size.
The objective function $\mathcal{L}_{\text{seg}}$ is the native training loss of SAM3, comprising the Intersection over Union (IoU) loss and cross-entropy loss.
Note that, with this continual searching strategy, the text prompt $T$ is no longer necessary.

\subsection{Hybrid Uncertainty Estimation for Sample Identification}
\label{sec:HUE}
To efficiently distinguish hard samples requiring subsequent human intervention from easy samples with already satisfactory annotation quality, we introduce an automatically calculated hybrid uncertainty metric, denoted as $U_{\text{HU}}(\cdot)$.
The metric quantifies the model's predicted map $y^{\text{pred}}$ (we omit indexes of iteration and sample subscripts for brevity) by integrating two complementary sub-metrics:
\begin{enumerate}
\item[(1)] \textbf{Prediction Entropy}.
We calculate the pixel-wise entropy of the predicted map to capture decision boundary ambiguity.
Let $p_i \in [0, 1]$ denote the value of the $i$-th pixel in $y^{\text{pred}}$, its Shannon entropy is defined as: $H(p_i) = -\big( p_i \log p_i + (1 - p_i) \log (1 - p_i) \big)$.
To strictly focus on informative and uncertain regions, we define a set of ambiguous pixels based on their local entropy thresholds: $\Omega = \{i \mid H(p_i) > 0.8\}$.
Consequently, the final prediction entropy for $y^{\text{pred}}$ is formulated as the mean entropy over this set: $U_{\text{ENT}}(y^{\text{pred}}) = \frac{1}{|\Omega|} \sum_{i \in \Omega} H(p_i)$.
\item[(2)] \textbf{Concept-Geometry Prompting Inconsistency}.
We transform the initial concept-driven predicted mask $y^{\text{pred}}$ into a bounding box.
The box serves as a geometric prompt for SAM3 to generate a geometry-driven mask $y^{\text{box}}$.
The uncertainty derived from this cross-prompt mechanism is defined by the IoU score between the two masks: $U_{\text{CGPI}}(y^{\text{pred}}) = 1 - \mathtt{IoU}(y^{\text{pred}}, y^{\text{box}})$.
\end{enumerate}
The final hybrid uncertainty score for a given predicted map $y^{\text{pred}}$ is formulated as the sum of these two sub-metrics:
$U_{\text{HU}}(y^{\text{pred}}) = U_{\text{ENT}}(y^{\text{pred}}) + U_{\text{CGPI}}(y^{\text{pred}})$.

After obtaining the hybrid uncertainties, we rank the predictions based on uncertainties and threshold them into hard and easy samples via a threshold $\tau$.
The top-$|\mathcal{D}_{t}^{\text{label}}|$ hard samples are routed for manual label correction, serving as the labeled samples $\mathcal{D}_{t}^{\text{label}}$ to support the continuous search for the concept embedding in the $(t+1)$-th iteration.
While the remaining hard samples are treated as the remaining unlabeled data and are fed into the next iteration.
Meanwhile, an optional early-exit strategy applied to the easy samples, completing their annotation process early and excluding them from subsequent iterations.

\section{Experiments}
\subsection{Experimental Setup}
\noindent \textbf{Baselines.}
We compare our M2C with six state-of-the-art (SOTA) few-shot segmentation (FSS) methods, categorized into training-based methods (Multiverseg \cite{wong2025multiverseg}, Tyche \cite{rakic2024tyche}) and training-free{\renewcommand{\thefootnote}{\arabic{footnote}}\setcounter{footnote}{0}\footnote{Here, `training-free' specifically means that the parameters of SAM3 remain unchanged and the approach does not require training by the FSS paradigm.} methods (SPFS-SAM \cite{wu2023self}, MAUP \cite{zhu2025maup}, Matcher \cite{liu2023matcher}, ProtoSAM \cite{ayzenberg2024protosam}) based on their requirement for FSS pre-training.

\noindent \textbf{Datasets.}
We conduct experiments on two medical datasets: Kvasir-SEG~\cite{jha2020kvasir} for endoscopic polyp segmentation and ISIC-2017~\cite{codella2018skin} for skin lesion segmentation.
These two datasets are both unseen during the pre-training of SAM3, which can effectively verify the model's zero-shot generalization capabilities.
For our evaluation protocol, we utilize the entire Kvasir-SEG dataset, comprising $1,000$ images.
Conversely, for the ISIC-2017 dataset, we strictly confine our evaluation to the official test set of $600$ images.
This restriction is implemented to prevent data leakage and to ensure a fair comparison with Tyche and Multiverseg, which were exposed to the ISIC training split during its pre-training.

\noindent \textbf{Evaluation Metrics.}
We employ Dice Similarity Coefficient to quantitatively evaluate segmentation performance.

\subsection{Implementation Details}
We implement our method based on the official pre-trained weight of SAM3~\cite{carion2025sam}.
To ensure a fair comparison, all SAM-based baselines (\textit{i.e.}, ProtoSAM, Matcher, SPFS-SAM, and MAUP) utilize SAM-H~\cite{kirillov2023segment}.
During the mask to concept continual searching, we freeze the entire SAM3 backbone and exclusively optimize the concept embedding.
The embedding optimization is executed on a single NVIDIA RTX 4090 GPU, optimized by AdamW~\cite{loshchilov2017decoupled} optimizer.
The searching step size $\eta$ is set to $5 \times 10^{-4}$ and follows a cosine decay strategy down to $5 \times 10^{-6}$. We set the uncertainty threshold $\tau$ to $0.6$ to distinguish between simple and hard samples. The concept embedding is initialized as an all-zero vector.

\subsection{Comparison in Few-shot Segmentation}
\label{sec:exp-fss}
\begin{table}[t]
\small
\centering
\caption{Quantitative comparison with SOTA FSS methods on Kvasir-SEG and ISIC-2017 datasets. We report the mean and standard deviation ($\mu \pm \sigma$) for the Dice metric as percentages(\%). Methods are divided into training-based (\textit{top}) and training-free (\textit{bottom}). The best results are highlighted in \textbf{bold}, and the second-best are \underline{underlined}.}
\vspace{-0.2cm}
\label{tab:sota}

\setlength{\tabcolsep}{4pt}
\renewcommand{\arraystretch}{0.9}

\resizebox{\textwidth}{!}
{%
\begin{tabular}{l ccc ccc}
\toprule
\multirow{2}{*}{\textbf{Methods}} & \multicolumn{3}{c}{\textbf{Kvasir-SEG}} & \multicolumn{3}{c}{\textbf{ISIC-2017}} \\
\cmidrule(r){2-4} \cmidrule(l){5-7}
 & 1-shot & 5-shot & 10-shot & 1-shot & 5-shot & 10-shot \\
\midrule

Tyche~\cite{rakic2024tyche}            & 26.9$\pm$8.2 & 43.1$\pm$2.0 & 49.9$\pm$1.5 & 37.2$\pm$15.9 & 68.1$\pm$3.5 & 69.7$\pm$2.8 \\
Multiverseg~\cite{wong2025multiverseg} & 35.8$\pm$6.7 & 48.5$\pm$4.6 & 50.5$\pm$3.1 & 46.3$\pm$5.1 & 70.8$\pm$3.5 & 71.6$\pm$2.3 \\
\midrule

ProtoSAM~\cite{ayzenberg2024protosam}  & \underline{72.1$\pm$4.8} & -- & -- & \underline{64.1$\pm$5.3} & -- & -- \\
Matcher~\cite{liu2023matcher}          & 67.1$\pm$13.5 & 68.3$\pm$1.9 & 69.2$\pm$4.9 & 54.2$\pm$11.3 & 58.8$\pm$8.2 & 58.3$\pm$2.3 \\
SPFS-SAM~\cite{wu2023self}             & 70.4$\pm$7.4 & \underline{78.5$\pm$1.1} & \underline{80.4$\pm$0.6} & 52.3$\pm$8.4 & \underline{72.3$\pm$2.5} & \underline{74.9$\pm$1.2} \\
MAUP~\cite{zhu2025maup}                & 46.9$\pm$5.7 & 60.6$\pm$3.9 & 58.3$\pm$6.3 & 53.0$\pm$8.5 & 55.7$\pm$6.6 & 59.3$\pm$4.4 \\
\textbf{M2C (Ours)}                    & \textbf{76.3$\pm$8.0} & \textbf{80.2$\pm$4.0}& \textbf{82.8$\pm$4.4} & \textbf{75.9$\pm$8.4} & \textbf{79.2$\pm$2.5} & \textbf{82.1$\pm$0.5} \\
\bottomrule
\end{tabular}%
}
\end{table}

\begin{figure}[t]
    \centering
    \includegraphics[width=\textwidth]{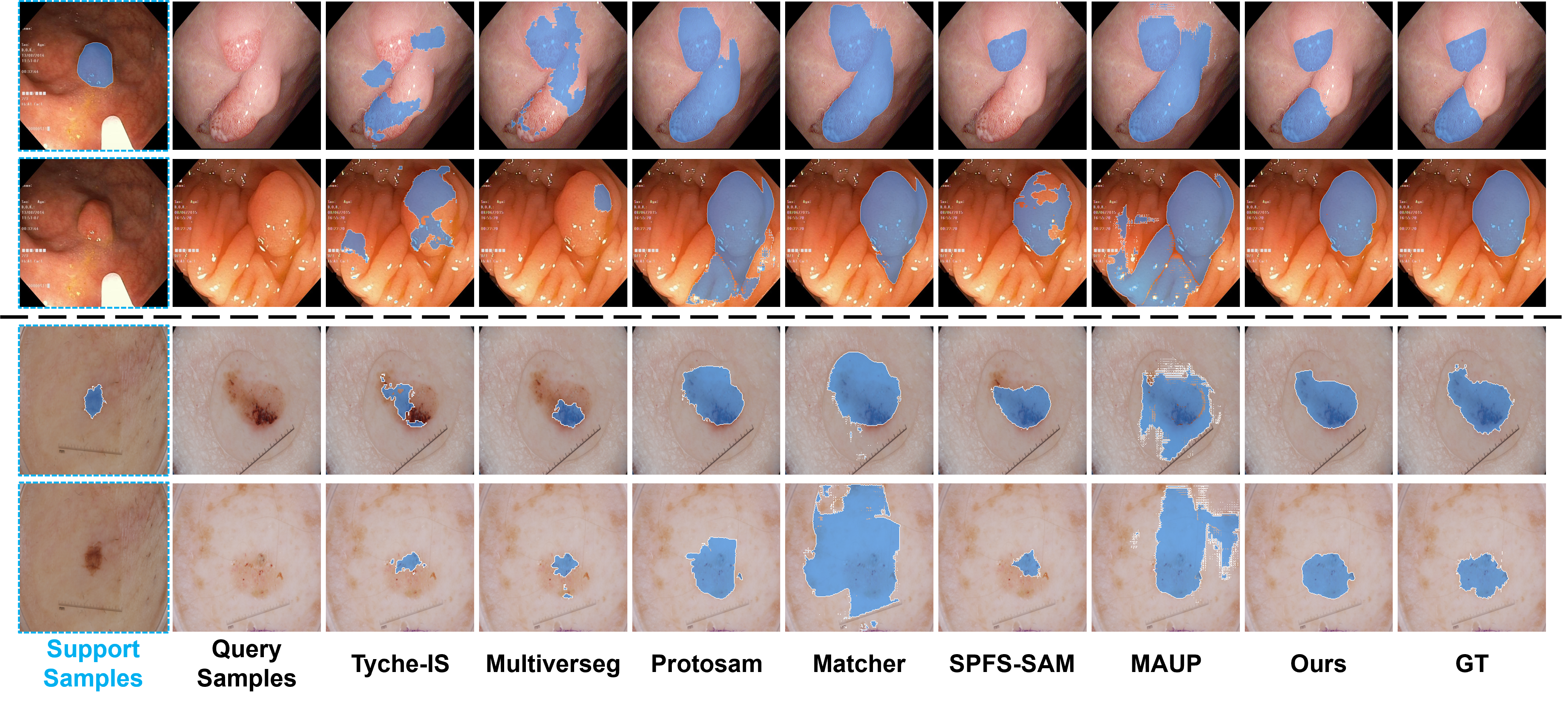} 
    \caption{Visualization of $1$-shot results on Kvasir-SEG (\textit{top}) and ISIC-2017 (\textit{bottom}).}
    \label{fig:qualitative}
\end{figure}
We compare our M2C with six FSS SOTAs on both the Kvasir-SEG and ISIC-2017 datasets independently, and all these methods are evaluated directly without further fine-tuning.
In our FSS evaluation protocol, the dataset is split into support and query sets using a $1:9$ ratio.
For $K$-shot segmentation, we test performance on the query set by forming episodes where one query sample is paired with $K$ randomly selected support samples.
To ensure statistical reliability, each query image undergoes five independent tests with different support sets, and the mean performance and standard deviation of these five rounds are reported.

\noindent \textbf{Quantitative Results.}
As shown in Table~\ref{tab:sota}, the training-based methods, Tyche and Multiverseg, typically exhibit limited performance. Despite undergoing FSS-paradigm pre-training on large-scale medical datasets, they still demonstrate poor generalization capabilities.
Conversely, our M2C significantly outperforms all competing methods across both datasets and under all shot settings. Particularly in the $1$-shot setting, M2C outperforms the second-best method, ProtoSAM, by a Dice improvement of $4.2\%$ ($76.3\%$ \textit{vs.} $72.1\%$) on Kvasir-SEG and $11.8\%$ ($75.9\%$ \textit{vs.} $64.1\%$) on ISIC-2017, respectively.
This superiority is attributed to our M2C mechanism, which can learn precise concept representations from extremely limited samples during test time, thereby facilitating robust and efficient adaptation to diverse scenarios.

\noindent \textbf{Qualitative Results.}
Fig.~\ref{fig:qualitative} visualizes the qualitative comparisons of four cases from the Kvasir-SEG and ISIC 2017 under the 1-shot setting.
As illustrated, our M2C can precisely segment ambiguous and complex boundaries of query images, even under significant visual variations.
In contrast, other training-free methods relying on SAM's geometric prompting mechanism are limited by the SAM's native capabilities, and the strategy of generating geometric prompts through feature matching struggles to overcome challenges such as blurry boundaries, background interference, and large cross-image visual variations.

\subsection{Comparison in Annotation Efficiency}
\label{sec:exp-annotation}
\begin{figure}[t]
    \centering
    \begin{minipage}{0.42\textwidth}
        \centering
        \includegraphics[width=\linewidth]{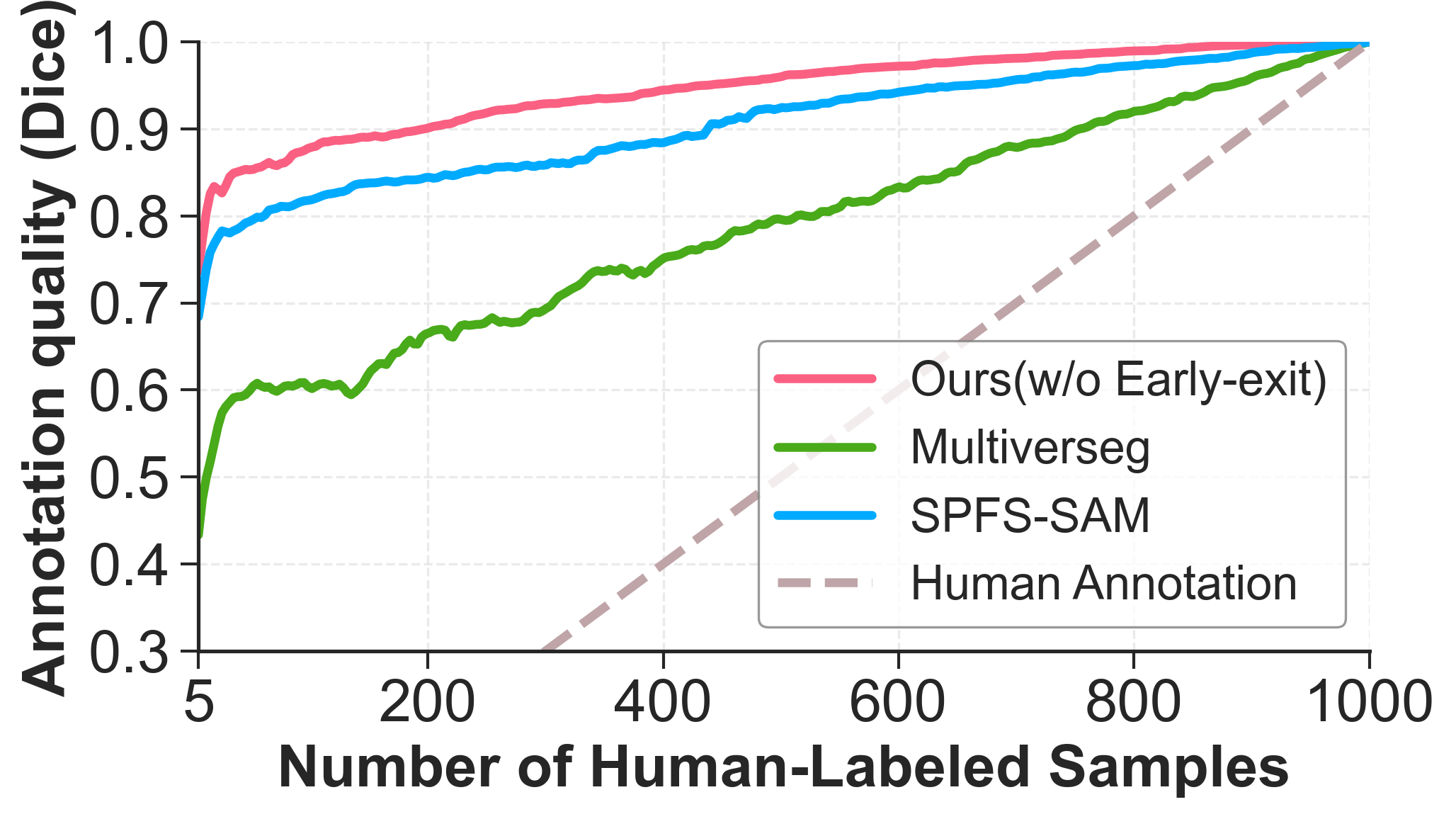}
        \centerline{(a) Kvasir-SEG}
    \end{minipage}\hfill
    \begin{minipage}{0.42\textwidth}
        \centering
        \includegraphics[width=\linewidth]{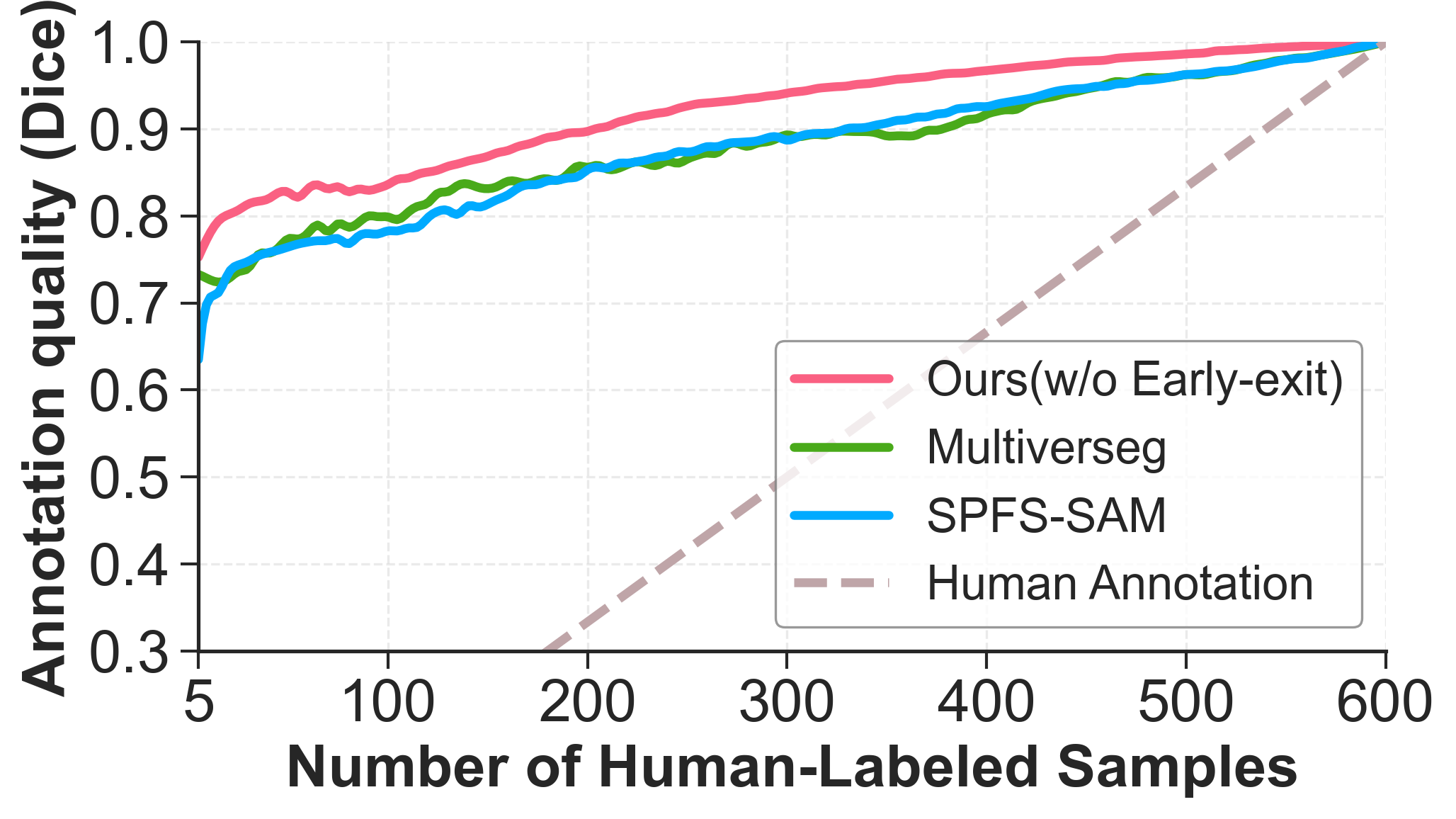}
        \centerline{(b) ISIC-2017}
    \end{minipage}
    \vspace{-0.2cm}
    \caption{Annotation efficiency curves on Kvasir-SEG and ISIC-2017 datasets.}
    \vspace{-0.5cm}
    \label{fig2}
\end{figure}
Motivated by the potential of these methods in automated data annotation, we compare the annotation efficiency of our M2C-based system with Multiverseg and SPFS-SAM.
We quantify this efficiency by measuring the overall dataset annotation quality under different numbers of human-labeled samples provided.
Specifically, the overall annotations include the human-labeled samples simulated using GT masks, and the remaining samples predicted by the model.
Notably, to ensure a fair comparison with methods lacking automatic annotation quality estimation mechanisms, we disable the early-exit strategy for easy samples.

In Fig.~\ref{fig2}, we plot the overall annotation quality curves as the number of annotated samples increases by intervals of $5$ (\textit{i.e.}, $|\mathcal{D}_{t}^{\text{label}}|=5$).
As observed, our method demonstrates the best efficiency, achieving satisfactory annotation quality with limited human annotation budget.
Furthermore, our method maintains a consistent superiority on different scenarios, showing the generalizability as an automated annotation tool and underscoring its significant practical value.

\subsection{Ablation Study}
\begin{figure}[t]
    \centering
    \makeatletter 
    \begin{minipage}[c]{0.48\textwidth}
        \centering
        \includegraphics[width=\linewidth]{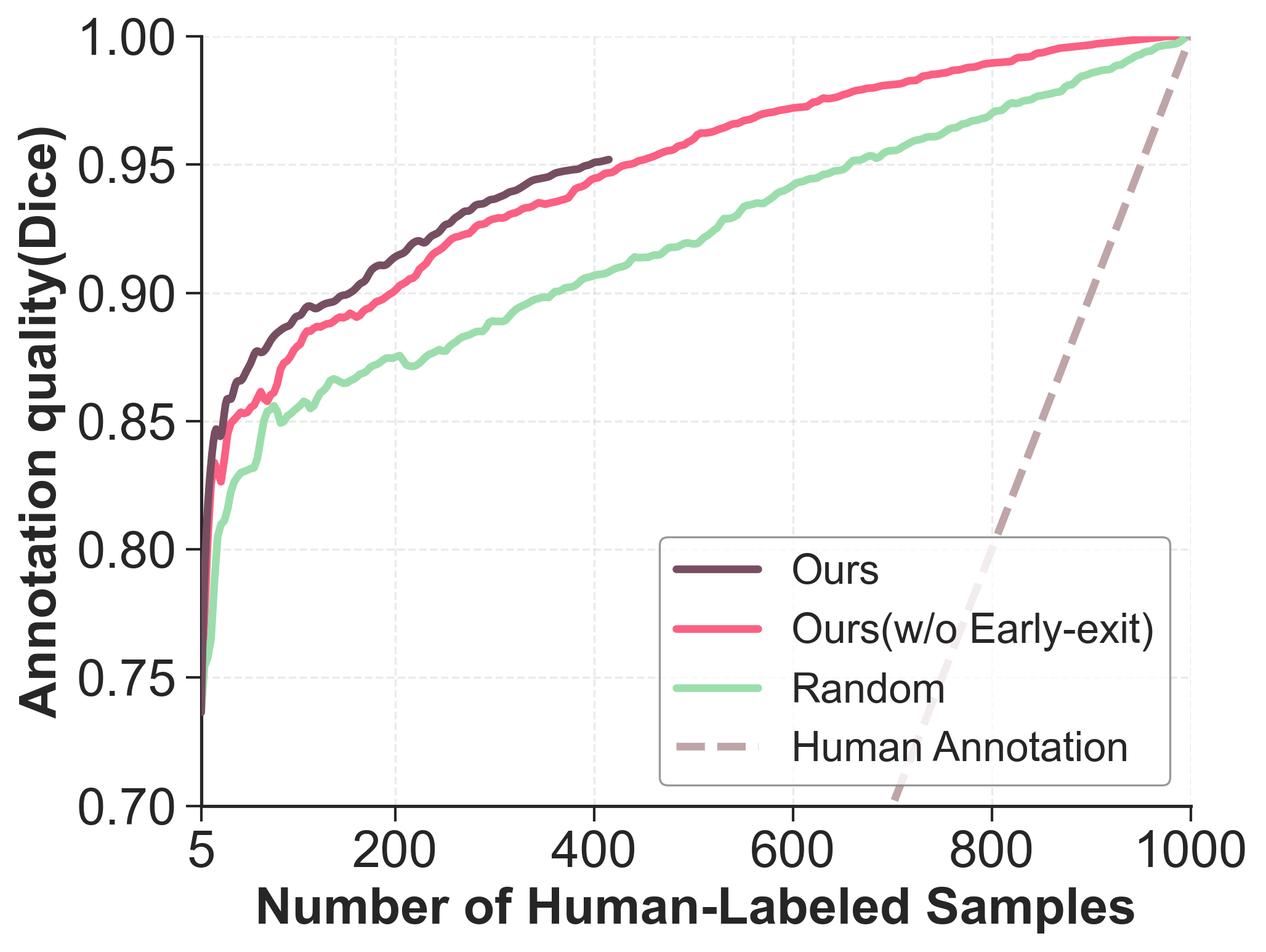}
        \vspace{-0.8cm}
        \caption{Ablation of HUE on Kvasir-SEG.}
        \label{fig:ablation}
    \end{minipage}\hfill
    \begin{minipage}[c]{0.48\textwidth}
        \centering
        \def\@captype{table}
        \vspace{-0.6cm}
        \caption{Ablation of two uncertainty sub-metrics ($U_{\text{ENT}}$, $U_{\text{CGPI}}$) on Kvasir-SEG. `Num' and `Ratio' represent the number and proportion of manually annotated samples required to achieve a Dice score of $90\%$, respectively, out of $1,000$ images.}
        \vspace{0.1cm}
        \label{tab:ablation_metrics}
        \setlength{\tabcolsep}{8pt}
        {
        \begin{tabular}{cc cc}
        \toprule
        \multicolumn{2}{c}{\textbf{Sub-metrics}} & \multicolumn{2}{c}{\textbf{Ann. Cost}} \\
        \cmidrule(lr){1-2} \cmidrule(lr){3-4}
        $U_{\text{ENT}}$ & $U_{\text{CGPI}}$ & \textbf{Num} & \textbf{Ratio} \\
        \midrule
        \checkmark &  & 345 & 34.5\% \\
         & \checkmark & 230 & 23.0\% \\
        \checkmark & \checkmark & \textbf{155} & \textbf{15.5\%} \\
        \bottomrule
        \end{tabular}%
        }
    \end{minipage}
    \vspace{-0.55cm}
    
    \makeatother 
\end{figure}
\noindent \textbf{Effect of HUE on Annotation Efficiency.}
To validate the efficacy of the proposed HUE, we conduct an ablation study on Kvasir-SEG comparing three configurations:
(1) \textit{Random}: We randomly select samples for manual annotation during each iteration.
(2) \textit{Ours (\textit{w/o} Early-exit)}: We use HUE to select hard samples and employ active manual annotation, but do not implement the early-exit strategy for easy samples.
(3) \textit{Ours}: Similar to \textit{Ours (\textit{w/o} Early-exit)} but implement the early-exit strategy for easy samples.
We plot the annotation efficiency curves for all configurations in Fig.~\ref{fig:ablation}.
\textit{Ours (w/o Early-exit)} surpasses the \textit{Random} configuration, verifying that our hybrid uncertainty metric can retrieve valuable samples to maximize dataset quality under fixed human annotation budgets.
Additionally, the complete configuration $\textit{Ours}$ achieves the most satisfactory annotation quality with minimal manual effort.
This demonstrates the vital role of early-exit for easy samples, ensuring they undergo precise segmentation through concept-aligned embeddings before continuous updates.

\noindent \textbf{Effect of Two Uncertainty Sub-metrics on Annotation Efficiency.}
Accurate uncertainty estimation is crucial for our human-in-the-loop annotation systems, as it directly dictates the early-exit of easy samples and the routing of hard samples for manual correction. In Table~\ref{tab:ablation_metrics}, we assess the impact of the two sub-metrics (prediction entropy $U_{\text{ENT}}$ and concept-geometry prompting inconsistency $U_{\text{CGPI}}$) on annotation efficiency by measuring the number of manually annotated samples required to achieve a target quality (Dice of $90\%$).
The results show relying solely on $U_{\text{ENT}}$ or $U_{\text{CGPI}}$ incurs a higher annotation cost than their combined strategy.
This demonstrates their complementarity and the importance of integrating them to accurately identify easy and hard samples.

\section{Conclusion}
In this work, we propose M2C, a novel mechanism designed to efficiently adapt SAM3 to specific medical imaging datasets under few-shot setting.
M2C can search the dataset-specific concept embedding from limited annotated masks, marking the first attempt to unlock the native textual concept prompting of SAM3 for medical scenarios without retraining.
Furthermore, we integrate M2C with HUE and human interaction to establish a semi-automatic, closed-loop annotation system.
Extensive experiments on the Kvasir-SEG and ISIC-2017 datasets demonstrate that M2C significantly outperforms SOTA FSS methods. Moreover, the proposed human-in-the-loop annotation system exhibits superior annotation efficiency and generalization, underscoring its potential as a highly efficient solution for real-world medical data annotation.

\begin{credits}
\subsubsection{\ackname}
This work was supported by National Natural Science Foundation of China (No.62401414 and No.62576145), Joint Funds of the Natural Science Foundation of Hubei Province, China (No.2026AFC0648), and research grants from Wuhan United Imaging Healthcare Surgical Technology Co., Ltd.


\subsubsection{\discintname}
The authors have no competing interests to declare that are relevant to the content of this article.
\end{credits}

\bibliographystyle{unsrt}  
\bibliography{references}

\end{document}